\documentclass[times, review, 10pt]{elsarticle}

\usepackage{amssymb}
\usepackage{amsmath}
\usepackage{adjustbox}
\usepackage{pifont}
\usepackage[pagebackref,breaklinks,colorlinks,allcolors=cvprblue]{hyperref}
\usepackage{booktabs,xcolor,cleveref}

\definecolor{cvprblue}{rgb}{0.21,0.49,0.74}

\definecolor{mygray}{gray}{0.9}
\definecolor{mygreen}{rgb}{0,.7,0}

\newcommand{\cmark}{\ding{51}}
\newcommand{\xmark}{\ding{55}}

\newcommand{\mypar}[1]{\noindent{\bfseries #1.}}

\journal{Pattern recognition}

\begin{document}

\begin{frontmatter}



\title{\textsc{Count2Density}: Crowd Density Estimation without Location-level Annotations} 


\author[label1]{Mattia Litrico}
\author[label2]{Feng Chen}
\author[label3]{Michael Pound}
\author[label2]{Sotirios A Tsaftaris}
\author[label1]{Sebastiano Battiato}
\author[label3]{Mario Valerio Giuffrida}


\affiliation[label1]{organization={University of Catania},
            city={Catania},
            country={Italy}}
\affiliation[label2]{organization={University of Edinburgh},
            city={Edinburgh},
            country={United Kingdom}}
\affiliation[label3]{organization={University of Nottingham},
            city={Nottingham},
            country={United Kingdom}}

\begin{abstract}
{Crowd density estimation is a well-known computer vision task aimed at estimating the density distribution of people in an image. The primary challenge in this domain is the reliance on fine-grained location-level annotations -- i.e., points placed on top of each individual—to train deep networks. Collecting such detailed annotations is both tedious, time-consuming, and poses a significant barrier to scalability for real-world applications. To alleviate this burden, we present \textsc{Count2Density}: a novel pipeline designed to predict meaningful density maps containing quantitative spatial information using \underline{only} count-level annotations (i.e., the total number of people) during training. To achieve this, \textsc{Count2Density} generates pseudo-density maps leveraging past predictions stored in a Historical Map Bank, thereby reducing confirmation bias. This bank is initialised using an unsupervised saliency estimator to provide an initial spatial prior and is iteratively updated with an Exponential Moving Average of predicted density maps. These pseudo-density maps are obtained by sampling locations from estimated crowd areas using a hypergeometric distribution, with the number of samplings determined by the count-level annotations. To further enhance the spatial awareness of the model and promote robust feature learning, we add a self-supervised contrastive spatial regulariser to encourage similar feature representations within crowded regions while maximising dissimilarity with background regions. Experimental results demonstrate that our approach significantly outperforms cross-domain adaptation methods and achieves better results than recent state-of-the-art approaches in semi-supervised settings across several datasets. Additional analyses validate the effectiveness of each individual component of our pipeline, including the self-supervised contrastive regulariser, confirming  the ability of \textsc{Count2Density} to effectively retrieve spatial information from count-level annotations and enabling accurate subregion counting.}
\end{abstract}


\end{frontmatter}

\section{Introduction}
\label{sec:intro}

Crowd density estimation aims to predict a spatial map representing the density of people in images \cite{lempitsky2010learning}. Despite considerable strides in this area, most current approaches require fine-grained location-level annotations to train a model, such as bounding boxes or points \cite{semiSupervised}. The need for large and fine-grained annotated datasets has become more pressing, as deep learning approaches for density estimation have gained popularity \cite{shanghaitech}.  However, collecting such annotations is tedious, error-prone, and time-consuming, often requiring domain-specific expertise, especially in scenarios where objects are difficult to detect \cite{jaiswal}. { This significant annotation cost and effort act as a major bottleneck for the widespread deployment of crowd density estimation in practical applications and for the development of large, diverse datasets required to train state-of-the-art deep learning models.}

To reduce the burden of collecting fine-grained location-level annotations, several approaches have been proposed, as summarised in \Cref{tab:fig1}. Semi-supervised methods predict density maps using datasets partially annotated with location-level data \cite{L2R}, reducing the number of annotated images but still requiring some location-level annotations. Cross-domain adaptation trains deep networks on fully annotated source datasets and adapts them to unlabelled target datasets, requiring location-level annotations for the source domain \cite{Fua}. Unsupervised methods avoid this requirement but perform significantly worse than semi-supervised and cross-domain approaches \cite{crowd_clip}. {While these methods offer advancements, they still fall short of fully eliminating the need for expensive location-level annotations or compromise on performance for practical applications.} Another approach is regression-based methods, whereby the model is trained to predict the total number of people in images \cite{xiong2024glance}. Although this approach helps to reduce the burden of collecting fine-grained annotations, its predictions do not provide spatial information, preventing subregion counting as these models do not generate any density maps. {Therefore, there remains a critical need for methods that can significantly reduce annotation costs by relying solely on readily available count-level data, while simultaneously retaining the ability to generate meaningful density maps that provide quantitative spatial information and enable fine-grained tasks such as subregion counting.}

A recent paper stated, ``\textit{Developing techniques to \underline{automatically} \underline{generate and refine labels} using unsupervised or semi-supervised learning approaches could contribute to model generalization and improve accuracy in real-world scenarios where annotated data may be scarce or unreliable.}'' \cite{jaiswal}. \textsc{Count2Density} aims to address this gap in the field.

To overcome the need for location-level annotations while still predicting spatial information for crowd density estimation, we propose \textsc{Count2Density}, a novel pipeline designed to directly predict meaningful density maps leveraging only count-level annotations during training, which are easier to collect \cite{semiSupervised}. To retrieve quantitative spatial information from counting values, we generate pseudo-density maps from past predictions stored in a historical map bank, updated with an exponential moving average of predicted density maps. At each iteration, an averaged density map is fetched from the bank and converted into an attention map. Then, we leverage the count-level annotation to sample as many locations as the total number of people in the image, using the attention map values as a probability prior. The generated pseudo density-map is then used to train the model in a \textit{self-supervised} fashion. To further improve the spatial awareness of the model, we incorporate a self-supervised contrastive spatial regulariser that encourages similar features within crowd regions, while promoting dissimilarity with background regions. { This unique integration of a Historical Map Bank and a contrastive spatial regulariser allows \textsc{Count2Density} to infer detailed spatial density information from  count-level supervision, a challenge previously under-explored in the literature.}

\begin{table*}[t]
\centering
\begin{adjustbox}{width=\textwidth}
\begin{tabular}{@{}llclc@{}}
\toprule
\textbf{Approach}             & \textbf{Annotations}  & \textbf{Annotations Cost}    & \textbf{Predictions}  & \textbf{Subregion counting}           \\ \midrule
Fully supervised \cite{BAY,MAN}    & {Location-level} & {High} & \color{mygreen}{Density maps} &  \textcolor{mygreen}{\cmark}   \\ 
\multicolumn{5}{l}{\textcolor{gray}{\small Note: Entire training set is annotated.}} \\ \midrule
Semi-supervised  \cite{
semisupervised2,GP}    & {Location-level} & \color{orange}{Medium} & \color{mygreen}{Density maps} &  \textcolor{mygreen}{\cmark}                                            \\ 
\multicolumn{5}{l}{\textcolor{gray}{\small Note: A portion of the training set is annotated.}} \\ \midrule
Cross-domain   \cite{BLA,Fua}      & {Location-level} & \color{orange}{Medium} & \color{mygreen}{Density maps} &  \textcolor{mygreen}{\cmark}                                 \\ 
\multicolumn{5}{l}{\textcolor{gray}{\small  Note: Source domain is fully annotated, target domain is unlabelled.}} \\ \midrule
Regression-based  \cite{wang2022joint,xiong2024glance}   & \color{mygreen}{Count-level}    & \color{mygreen}{Low} & {Crowd counts} &  \textcolor{red}{\xmark}   \\ 
\multicolumn{5}{l}{\textcolor{gray}{\small Note: They do not directly predict density maps, which are extracted from feature maps lacking quantitative density  information.}} \\ \midrule \midrule
\textbf{\textsc{Count2Density}} & \color{mygreen}{Count-level}    & \color{mygreen}{Low} & \color{mygreen}{Density maps} &  \textcolor{mygreen}{\cmark}  \\ 
\multicolumn{5}{l}{\textcolor{gray}{\small Note: Count-level annotations are used to generate pseudo-density maps.}} \\
\bottomrule
\end{tabular}
\end{adjustbox}
\caption{Differences with previous approaches. \textsc{Count2Density} is trained to predict density maps, by generating pseudo-density maps from count-level annotations. This way of training reduces the effort of collecting location-level annotation while enabling subregion counting.}
\label{tab:fig1}
\end{table*}

We benchmark our method on several major crowd counting datasets \cite{shanghaitech,ucf_qnrf,jhu-crowd++} outperforming the state-of-the-art by a large margin. {When compared to cross-domain approaches, our method reduces the Mean Absolute Error (\textsc{mae}) by $49.1$ (from $198.3$ to $149.2$) on \textsc{ucf-qnrf}, and by $7.8$ (from $99.3$ to $91.5$) on \textsc{ShanghaiTech-a}.}
We also present results obtained by training our method in a semi-supervised manner, further outperforming related state-of-the-art methods. Qualitative results demonstrate  that \textsc{Count2Density} generates meaningful density maps from count-level annotations, enabling subregion counting. Additional analyses validate  the effectiveness of each individual component in our pipeline.

The main contributions of our work are as follows:

\begin{itemize}
    
    \item To reduce the cost of collecting fine-grained location-level annotations, we introduce \textsc{Count2Density}, a novel pipeline that predicts meaningful density maps using only count-level annotations, { overcoming the fundamental limitation of regression-based methods that lose  spatial information.}

    \item To capture spatial information, \textsc{Count2Density}  leverages count-level annotations to generate pseudo-density maps by sampling points from past predictions. Additionally, we add a self-supervised contrastive spatial regulariser to enhance the model's spatial awareness, guiding representations in both crowd and background areas. { These two methodological contributions are central to the ability of our model to infer rich spatial information from count-level annotations.}
    
    \item We evaluate \textsc{Count2Density}  on several benchmark datasets, achieving significant improvements over state-of-the-art cross-domain and semi-supervised approaches. Our analyses demonstrate the ability of \textsc{Count2Density}  to effectively retrieve spatial information from count-level annotations { and establish a new paradigm for density estimation without the need for burdensome and costly annotations}.

\end{itemize}

\section{Related Works}
\mypar{Crowd density estimation} Density estimation has achieved great success in crowd counting due to its robustness in handling crowded scenes and complex backgrounds. From a computer vision perspective, perspective distortion, caused by varying filming angles and distances, poses a significant challenge. Some research has addressed this distortion by using perspective maps for additional supervision \cite{yan2021crowd}, while other studies have proposed perspective-map-free approaches \cite{zhang2016single}. In \cite{zhang2016single}, a multi-column network with different convolutional kernel sizes in each branch of the model was introduced to capture multi-scale information. In \cite{liu2019context}, the authors proposed a context-aware model to adaptively handle scale variations. Recent works \cite{Yang2022CrowdFormerAO} also leverage transformer-based models to capture global relationships in crowd counting. More recently, \textit{CrowdDiff} was proposed, where a diffusion model is trained to generate multiple hypotheses of crowd density \cite{CrowdDiff}. These approaches require location-level supervision during training. Given the difficulty and cost of collecting such annotations, other methods have been developed to learn from limited supervision.

\mypar{Learning from limited location-level annotations} Unsupervised strategies for crowd density estimation have been proposed, but they generally yield unsatisfactory performance \cite{crowd_clip}. Semi-supervised approaches, on the other hand, have achieved better results. \textsc{l2r} \cite{L2R} leverages the abundant availability of unlabelled crowd images by learning to rank. \textsc{matt} \cite{semiSupervised} uses both location-level and count-level annotations, promoting consistency between predictions generated by multiple auxiliary tasks. \textsc{irast} \cite{semisupervised2} is a self-training method that incorporates inter-relationships between different predictors to produce reliable pseudo-labels for semi-supervised learning. \textsc{gp} \cite{GP} is another semi-supervised crowd counting method that estimates pseudo-labels through a Gaussian Process-based iterative learning mechanism. \textsc{ac-al} \cite{active_learning} introduces an active learning framework to gradually collect point annotations. A different approach is taken by PAL \cite{partialAnn}, where a deep neural network for density estimation is trained with partial annotations, requiring only a small, annotated region in each image. While semi-supervised methods reduce the reliance on extensive location-level annotations, they still necessitate a subset of location-level labels to produce meaningful density maps. Another way to alleviate the need for location-level annotations is through cross-domain learning, such as domain adaptation \cite{BLA}, where a model is trained on a fully annotated source domain and then adapted to an unlabelled target domain. Unlike these methods, \textsc{Count2Density} generates meaningful density maps without requiring any location-level annotations.

\mypar{Learning from count-level annotations} Approaches that rely solely on count-level annotations are typically categorised as weakly supervised methods \cite{xiong2024glance,Litrico2021}. These methods are primarily regression-based, meaning they predict the total number of people in a scene (\textit{c.f.} \Cref{tab:fig1}). Given their significance in the field, it is important to review the literature on regression-based approaches. In \cite{regression_counting}, the authors propose a regression-based weakly supervised method that uses a sorting network to improve crowd counting. In \cite{kumagai2017mixture}, multiple specialized CNNs are used, which are adaptively selected based on the appearance of an image through a gating network. In \cite{xiong2024glance}, the authors introduce a strategy for crowd counting using a large number of ranking labels and a few images annotated with count labels. Transformer-based weakly supervised approaches have also been proposed, such as \cite{liang2022transcrowd}. Although these methods predict accurate crowd counts, they infer density maps from feature activations. However, this approach to estimating density maps loses quantitative spatial information, as feature map values exhibit different statistical characteristics than density values, which impedes subregion counting. In contrast, our method leverages only count-level annotations and directly predicts meaningful density maps, where values are correlated with crowd density. This also enables subregion counting by integrating over areas of interest. In our experiments, we do not compare with regression-based methods, as their output (counts) differs from ours (densities), as shown in \Cref{tab:fig1}.

\mypar{Summary} Our approach differs from the state-of-the-art in crowd density estimation in terms of how the model is trained. \textsc{Count2Density} is not a regression model, as it directly outputs density maps. Count-level annotations are used to generate pseudo-density maps, rather than serving as a direct form of supervision. Training on pseudo-density maps allows us to retrieve quantitative spatial information from count-level annotations, enabling effective crowd counting in both the entire image and subregions.

\section{Proposed Method}
\begin{figure*}[t]
    \centering
    \includegraphics[width=\linewidth]{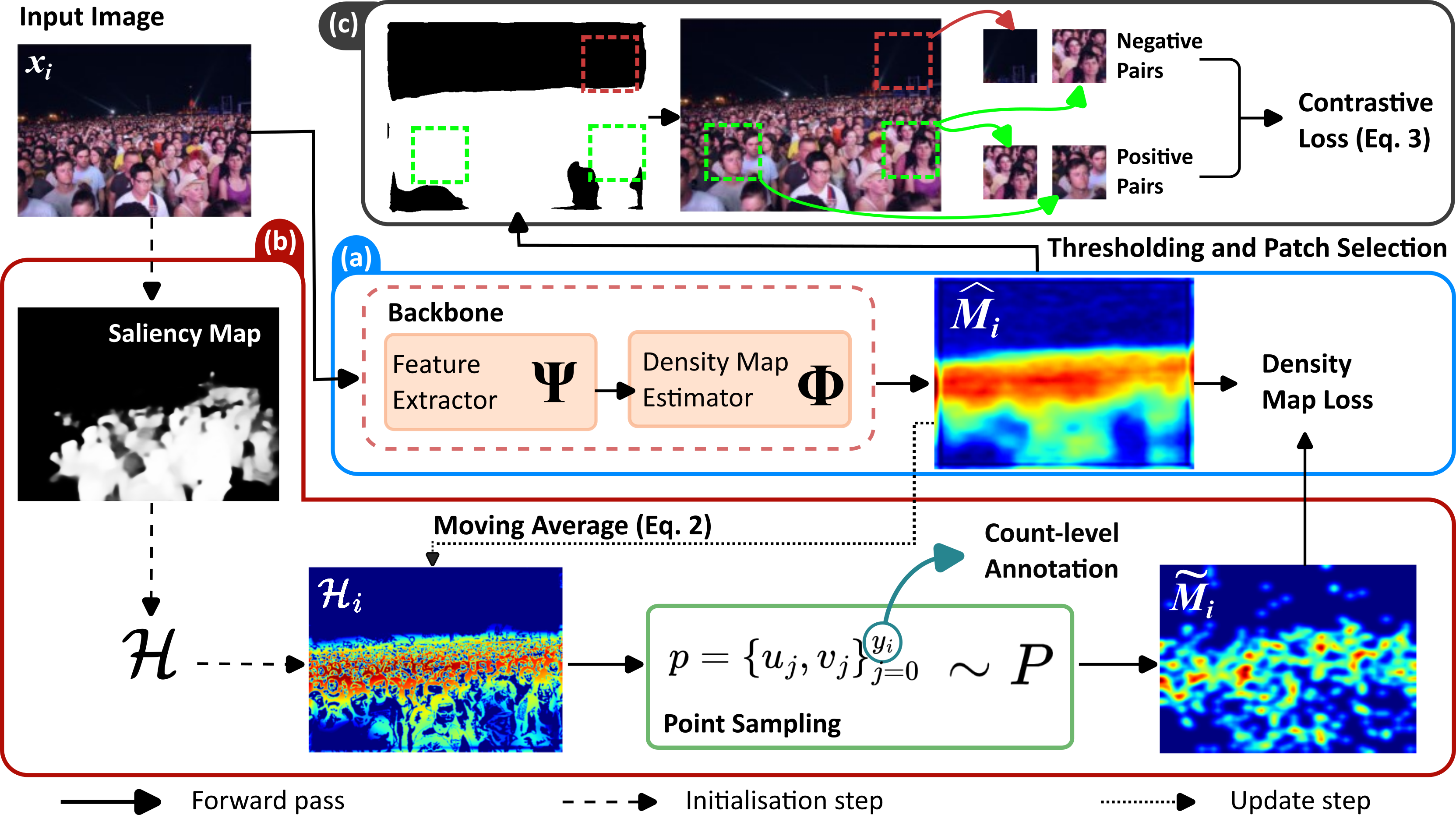}
    \caption{Overview of \textsc{Count2Density}. (a) The input image $x_i$ is provided to the backbone to predict a density map $\hat{M}_i$. (b) $x_i$ is provided to an unsupervised saliency estimator to initialise the Historical Map Bank $\mathcal{H}$ (\Cref{sec:unsupervised_noisy_spatial_prior}). The historical map bank is updated at each epoch, using an Exponential Moving Average (\Cref{eq:moving_avg}) of the predicted density map $\hat{M}_i$. For each image $x_i$, the information in $\mathcal{H}$ is retrieved and $\mathcal{H}_i$ is used to sample $y_i$ (count-level annotation) points to generate a pseudo-density map $\tilde{M}_i$ that is used as source of self-supervision to update the weights of the feature extractor $\Psi$ and the density map estimator $\Phi$. (c) Using the threshold image obtained from the predicted density map $\tilde{M}_i$, we distinguish between crowded and background regions. Positive pairs are then formed by pairing crowded areas.  Note that the patch selection is performed on the feature maps extracted by $\Psi$. (Best viewed in colour.)  }
    \label{fig:our_method}
\end{figure*}

\noindent  \Cref{fig:our_method} offers an overview of our approach: an input image $x_i$ is provided into the model to predict a density map. The predicted density map $\hat{M}_i$ is used in the loss function, minimising the difference with a generated pseudo-density map $\tilde{M}_i$. The pseudo-density map is obtained by sampling as many locations as in the ground-truth count-level annotation, leveraging past predictions stored in a \textit{Historical Map Bank} as probability prior. To mitigate the confirmation bias, the historical map bank is updated iteratively using an exponential moving average. Lastly, to improve the spatial awareness of the model, we adopt a self-supervised contrastive loss to encourage distinguishable features between crowd and background regions. 

\subsection{Problem Statement}
\label{sec:problemsetup}

\noindent Let $\mathcal{D} = \{x_i, y_i\}_{i=0}^N$ be a dataset containing $N$ images $x_i\in \mathbb{R}^{3\times H \times W}$ and count-level labels $y_i \in \mathbb{N}$. Given a feature extractor $\Psi(x_i)$, which extracts a feature vector $z_i\in \mathbb{R}^D$ from an input image $x_i$, and a decoder $\Phi$, we train the network to predict a density map $\hat{M}_i \in \mathbb{R}^{\hat{H} \times \hat{W}}$ obtained as $\hat{M}_i = \Phi(z_i)$. The predicted total count $\hat{y}$ is calculated by integration from $\hat{M}_i$ as described in \cite{lempitsky2010learning}:\footnote{Note that \cref{eq:integral} can be adopted for subregion counting.}

\begin{equation}
\label{eq:integral}
\hat{y}_i = \sum_{u=1}^{W} \sum_{v=1}^{H} \hat{M}_i(u,v).
\end{equation}


\subsection{Generating Pseudo-Density Maps}
\label{sec:unsupervised_noisy_spatial_prior}

\noindent Count-level annotations lack quantitative spatial information. The purpose of \textsc{Count2Density} is to retrieve spatial information from only count-level annotations, as count values are \textit{the only source of supervision} used during training. Inspired by the paradigm of training from noisy labels,  \textsc{Count2Density} generates pseudo-density maps from count values by sampling locations that are likely to belong to crowd-dense regions. Any sampling errors are treated as noise within the pseudo-labels. The generated pseudo-density map retrieves quantitative spatial information from count-level annotations, and these are used as self-supervision for training the model. 

\Cref{fig:our_method}(b) shows how the pseudo-density maps are generated during training. We build a \textit{historical map bank} $\mathcal{H} \in \mathbb{R}^{N \times H \times W}$ that stores past predictions $\hat{M_i}$ for each input image $x_i$ in the training set. Specifically, each entry $\mathcal{H}_i \in \mathbb{R}^{H \times W}$ in the historical map bank is updated at each epoch $t$ by calculating an exponential moving average of the predicted density map $\hat{M}_i$ as follows:

\begin{equation}
    \label{eq:moving_avg}
    \mathcal{H}_i^{(t)} = \alpha \hat{M}_i^{(t)} + (1-\alpha)\mathcal{H}^{(t-1)}_i,
\end{equation}

\noindent where $0\leq\alpha\leq1$ is a hyperparameter controlling the update of the historical map bank, and the subscript $i$ indicates that the $i$-th map is obtained by averaging past predictions from the $i$-th training image. In this way, the averaged density maps in the historical map bank act as an ensemble of networks, alleviating confirmation bias \cite{DivideMix}. Note that for each image in the training set, we store a corresponding map in the bank. This means that the output from the $i$-th image, contribute in updating only the i-th map in the bank.

The historical map bank calculated in \Cref{eq:moving_avg} is computed for each image in the training set and used as a probability prior to generate the pseudo-density maps at each training iteration. Specifically, we take the $i$-th map in the bank $\mathcal{H}_i^{(t)}$ at epoch $t$ and we normalise\footnote{The normalisation is performed by dividing each pixel by the maximum value in the map $\mathcal{H}_i^{(t)}$. After that, we apply a Gaussian filter to smooth out the result.} it to make its values represent the probability of a pixel being in a crowd region. We denote the normalised map as $\widehat{\mathcal{H}_i^{(t)}}$. Then, to generate the pseudo-density maps, we sample
{$p = \left\lbrace u_j,v_j\right\rbrace^{y_i}_{j=0}$} number of locations equal to the count label $y_i$, { where $u_j$ and $v_j$ indicate the horizontal and vertical coordinates of the sampled point, respectively}. The sampling is performed using a hypergeometric distribution $\mathcal{M}(y_i, \widehat{\mathcal{H}_i^{(t)}})$, where $y_i$ is the number of sampled locations and $\widehat{\mathcal{H}_i^{(t)}}$ is the probability of each pixel being sampled. This process is equivalent to performing $y_i$ Bernoulli samplings without replacement.

At the beginning of the training, the historical map bank contains arbitrary values, diminishing the spatial information in the generated pseudo-density maps. To address this issue, we opt to initialise the historical map bank using an unsupervised saliency estimator. This estimator predicts a binary map, where non-zero pixels indicate salient regions within an image. This initialisation provides a noisy prior for generating subsequent pseudo-density maps. An example of the saliency estimation is depicted in \Cref{fig:our_method}.
Unsupervised saliency estimation methods \cite{saliency_estimation11} have been shown to perform on par with their supervised counterparts \cite{saliency_estimation18} across several scenarios. The current literature is rich in saliency estimation approaches, using hand-crafted priors or relying on videos to learn a salient object detector. Here, we use \textsc{bas-net} \cite{basnet}, as it is fully \textit{unsupervised} and exhibits good performance in different scenarios.
{To initialise the Historical Map Bank, we run \textsc{bas-net} on each of the training images and we use the prediction to initialise the corresponding map in the bank. More formally, we initialise the Historical Map Bank as follow:
$$\mathcal{H}_i^{0} = \operatorname{BASNET}(x_i),$$
where $x_i$ is a training image and $\mathcal{H}_i^{0}$ is the corresponding map in the Historical Map Bank. The superscript $0$ means that this initialisation is performed before the training ($t=0$). }
In \Cref{sec:analysis}, we present an analysis of the performance achieved by comparing various saliency estimators. We also show the performance of \textsc{Count2Density} without saliency initialisation. {Our analyses will show that initialising $\mathcal{H}$ with \textsc{bas-net} \cite{basnet} improves performance, and it does not require any extra source of supervision, as it was designed to operate without ground-truth.}

\subsection{Learning Spatially Consistent Features}
\label{sec:contr_loss}


\noindent Crowd images can be divided into crowded and background areas. Specifically, crowd regions are typically characterised by distinct patterns, making them significantly different from background areas (\textit{e.g.}, sky, buildings, etc.). Based on this observation, we leverage a self-supervised contrastive feature regulariser to promote spatial consistency for the feature extractor $\Psi(\cdot)$, as shown in \Cref{fig:our_method}(c). This regulariser encourages the feature distribution within crowded regions to be similar. The distance between representations extracted from crowded areas is minimised, while the distance between features extracted from background areas is maximised. Contrastive learning \cite{contrastive_loss} requires positive pairs $(f, f^+)$ and negative pairs $\{(f, f_0^-), (f, f_1^-), ..., (f, f_K^-)\}$, which we select using the predicted density map $\hat{M}$. Specifically, we normalise $\hat{M}$ and then apply a threshold to select non-zero pixels that corrispond to crowd regions. At this point, we build positive pairs by matching two crowded areas and negative pairs by matching a crowd and a background area. We then optimise the following contrastive loss:

\begin{equation}
\label{eq:contr_loss}
\mathcal{L}^{ctr} = - \log \frac{\exp\left(\Psi(f)^T \cdot \Psi(f^+)/\tau\right)}{ \sum_{k=0}^K \exp\left(\Psi(f)^T \cdot \Psi(f_k^-)/\tau\right)} ,
\end{equation}

\noindent where the temperature $\tau$ controls the scale of the dot product, and $T$ indicates the transposition operation.

\subsection{Total Objective Function}
\label{sec:obj_fun}
\noindent We train \textsc{Count2Density} by optimising the following combined objective function:
\begin{equation}
    \label{eq:loss}
    \mathcal{L}(x_i;\Psi,\Phi) = \mathcal{L}^{map} + \mathcal{L}^{ctr},
\end{equation}

\noindent where $\mathcal{L}^{map}$ and $\mathcal{L}^{ctr}$ are equally weighted.

Our approach is agnostic to the choices of $\Psi$ and $\Phi$ (the backbone), and we adapted several state-of-the-art architectures to demonstrate its effectiveness. As different backbones use different loss functions for the $\mathcal{L}^{map}$ term in \cref{eq:loss}, we modify $\mathcal{L}^{map}$ accordingly.

\section{Experimental Results}

\mypar{Implementation Details} We used PyTorch as the deep learning framework. Since \textsc{Count2Density} is agnostic to the backbone  (\textit{c.f.} \Cref{sec:obj_fun}), we integrated our training framework into the following state-of-the-art architectures: (i) \cite{NCC}, which models noisy annotations for crowd counting (referred to as \textsc{ncc}); (ii) \cite{BAY}, which adopts a Bayesian loss (\textsc{bl}); (iii) \cite{GL}, which employs a generalised loss (\textsc{gl}); and lastly, (iv) the Multifaceted Attention Network \cite{MAN} (\textsc{man}). {To integrate our framework with these methodologies, we use their same backbones and losses ($\mathcal{L}^{map}$) Moreover, we also evaluate our framework using a standard U-Net architecture and a Mean-Squared Error (MSE) as $\mathcal{L}^{map}$.}   For fair comparisons, hyperparameters were set as in their respective implementations. We set $\alpha=0.7$ in \cref{eq:moving_avg}, and $\tau=0.07$ in \cref{eq:contr_loss}. The value of $\alpha$ was chosen following a grid search, while the value of $\tau$ was taken from \cite{litrico_cvpr}.

\mypar{Datasets} We adopted the following datasets: \textsc{ShanghaiTech} \cite{shanghaitech}, \textsc{ucf-qnrf} \cite{ucf_qnrf}, \textsc{jhu-crowd++} \cite{jhu-crowd++}, \textsc{nwpu-crowd} \cite{nwpu}. \textsc{ShanghaiTech} consists of two subsets: Part A (482/300 images for training/testing) and Part B (716/400). \textsc{ucf-qnrf} is a large-scale dataset with 1,535 high-resolution images (1,201/334 for training/testing). \textsc{jhu-crowd++} contains 4,317 images (2,722/500/1,600 for training/validation/testing). \textsc{NWPU-Crowd} \cite{nwpu} is a massive crowd counting dataset containing highly congested crowds and appearance variations.

\mypar{Evaluation Metrics} As in \cite{BAY, MAN,NCC,GL}, we use the \textit{Mean Absolute Error} (\textsc{mae}) and the \textit{Root Mean Squared Error} (\textsc{mse}) as evaluation metrics: 
\begin{equation}
    \label{eq:metrics}
    \operatorname{MAE} = \frac{1}{N} \sum_{i=1}^N \left|y_i - \hat{y}_i\right|, \operatorname{MSE} = \left(\frac{1}{N} \sum_{i=1}^N \left(y_i - \hat{y}_i\right)^2\right)^{\frac{1}{2}},
\end{equation}
where $N$ is the number of images, while $y_i$ and $\hat{y}_i$ are the count-level annotations and predicted counts (\textit{c.f.} \cref{eq:integral}), respectively. For a qualitative evaluation of the predicted density maps, we use the \textit{Structural Similarity Index Measure} (\textsc{ssim}) and \textit{Peak Signal-to-Noise Ratio} (\textsc{psnr}).

\begin{table*}[t!]

\begin{adjustbox}{width=\textwidth}
\begin{tabular}{l|ccc|cc|cc|cc|cc|cc}
\toprule
 &  \multicolumn{3}{c|}{\textbf{Annotation Type}} & \multicolumn{2}{c}{\textsc{ucf-qnrf}\cite{ucf_qnrf}} & \multicolumn{2}{c}{\textsc{ShT-A}\cite{shanghaitech}} & \multicolumn{2}{c}{\textsc{ShT-B}\cite{shanghaitech}} & \multicolumn{2}{c}{\textsc{jhu-crowd++}\cite{jhu-crowd++}} & \multicolumn{2}{c}{\textsc{nwpu-crowd}}\cite{nwpu}\\ \midrule 
\textbf{Method} & \textbf{Locations} & \textbf{Counts} & \textbf{Self-labels} & \textsc{mae} & \textsc{mse} & \textsc{mae} &\textsc{mse} & \textsc{mae} & \textsc{mse} & \textsc{mae} & \textsc{mse} & \textsc{mae} & \textsc{mse} \\
\midrule
\multicolumn{13}{l}{\textit{\textbf{Density estimation Methods}}}  \\ \midrule
\multicolumn{13}{l}{\textit{\textbf{Fully Supervised}}}  \\ 
 \textsc{ncc} \cite{NCC}  & \cmark & & &  85.8 & 150.6 & 61.9 & 99.6 & \underline{7.4} & \textbf{11.3} & 67.7 & 258.3 & 96.9 & 534.2  \\
 \textsc{gl} \cite{GL}    & \cmark & & &   \underline{84.3} & \underline{147.5} & \underline{61.3} & \underline{95.4} & \textbf{7.3} & \underline{11.7} & 59.9 & 259.5 & \underline{79.3} & \underline{346.1} \\
 \textsc{bl} \cite{BAY} & \cmark & & &   85.8 & 150.6 & 61.9 & 99.6 & \underline{7.4} & \textbf{11.3} & 67.7 & 258.5 & 105.4 & 454.2  \\
 \textsc{man} \cite{MAN}   & \cmark & & &   \textbf{77.3} & \textbf{131.5} & \textbf{56.8} & \textbf{90.3}  & -- & -- & \textbf{53.4} & \textbf{209.9} & \textbf{76.5} & \textbf{323.0}  \\ \midrule \midrule

\multicolumn{14}{l}{\textit{\textbf{Semi-supervised }}} \\ 

 \textsc{l2r} \cite{L2R}  & $\blacktriangle$ & & \cmark &  148.9 & 249.8 & 90.3 & 153.5 & 15.6 & 24.4 & -- & -- &  125.0 & 501.9 \\
 \textsc{uda} \cite{UDA}  & $\blacktriangle$ & & & -- & -- &  93.8 & 157.2 & 15.7 & 24.1 & -- & -- & -- & -- \\
 \textsc{mt} \cite{MT}  & $\blacktriangle$ & & & -- & -- &  94.5 & 156.1 & 15.6 & 24.5 & -- & -- &  129.8 & 515.0 \\
 \textsc{ict} \cite{ICT} & $\blacktriangle$ & & & -- & -- & 92.5 & 156.8 & 15.4 & 23.8 & -- & -- & -- & -- \\
 \textsc{matt} \cite{semiSupervised}  & $\blacktriangle$ & \cmark & &  -- & -- & {80.1} & {129.4} & 11.7 & 17.5 & -- & -- & -- & --\\
 \textsc{irast} \cite{semisupervised2}  & $\blacktriangle$ &  & \cmark & {135.6} & {233.4} & 86.9 & 148.9 & 14.7 & 22.9 & {135.0} \textdagger & {450.7} \textdagger& {122.1} \textdagger & {484.3} \textdagger\\
 \textsc{gp} \cite{GP}  & $\blacktriangle$ & & &  160.0 & 275.0 & 102.0 & 172.0 & 15.7 & 27.9 & -- & -- & -- & --\\
 \textsc{pal} \cite{partialAnn} & $\blacktriangle$ & & & 128.1 & 218.0 & 72.7 & {111.6} & 12.0 & 18.7 & 129.6 & 400.4 & 178.7 & 1080.4\\
 \textsc{ac-al} \cite{active_learning}  & $\blacktriangle$ & & & -- & -- & 87.9 & 139.5 & 13.9 & 26.2 & -- & -- & -- & --\\
\textsc{cu} \cite{CU} &  $\blacktriangle$ & & & \underline{104.0} & \underline{164.2} & {70.7} & 116.6 & \textbf{9.7} & \textbf{17.7} & {74.8} & {281.6} &  {108.7} & {458.0}\\
 \textbf{\textsc{Count2Density} (Semi-sup.) 5\%}  & $\blacktriangle$ & \cmark &  & 120.1 & 201.0 & \underline{70.5} & \underline{105.6} & 13.0 & 20.1 & \underline{65.8} & \underline{189.1}  & \underline{107.5} & \underline{448.1} \\
{\textbf{\textsc{Count2Density} (Semi-sup.) 10\%}}  & $\blacktriangle$ & \cmark &  & \textbf{102.2} & \textbf{152.4} & \textbf{62.7} & \textbf{98.3} & \underline{10.2} & \underline{16.7} & \textbf{60.9} & \textbf{181.5}  & \textbf{99.7} & \textbf{398.0} \\
\midrule \midrule 

\multicolumn{14}{l}{\textit{\textbf{Cross-Domain}}}\\ 

\textsc{se+fd} \cite{SE+FD}  & * & & & -- & -- &  129.3 & 187.6 & 16.9 & 24.7 & -- & -- & -- & -- \\
\textsc{cd-cc} \cite{Fua}  & * & &  & 198.3 & 332.9 & 109.2 & 168.1 & \textbf{11.4} & \textbf{17.3} & -- & -- & -- & --\\
\textsc{bla} \cite{BLA}  & * & &  & 198.9 & 316.1 & 99.3 & 145.0 & \underline{11.9} & \underline{18.9} & -- & -- & -- & --\\
\midrule 


\textsc{irast} \cite{semisupervised2} \textdagger &  & \cmark & &  437.1 & 498.4 & 218.5 & 289.9 & 64.3 & 69.9 & 248.4 & 683.5 & 287.3 & 824.1\\
{Yang et al. \cite{regression_counting}} &  & \cmark & & - & - & {104.6} & {145.2} & {12.3} & {21.2} & - & - & - & - \\
\textbf{\textsc{Count2Density}}  & & \cmark &  & 191.2 & 321.3 & 95.3 & 154.9 & 15.4 & 23.7 & 115.8 & 385.7   & 132.3 & 514.8\\
\textbf{\textsc{Count2Density} (\textsc{ncc})} & & \cmark &  &  164.6 & 270.5 & 100.3 & 165.2 & 35.1 & 47.3 & 102.2 & 298.2 & \underline{122.9} & \underline{489.9}  \\
\textbf{\textsc{Count2Density} (\textsc{bl})}   & & \cmark & &  162.4 & 282.2 & 97.5 & 149.8 & 50.8 & 79.4 & \underline{92.4} & \underline{269.4}  & 125.7 & 495.4 \\
\textbf{\textsc{Count2Density} (\textsc{gl})} & & \cmark & &  \underline{159.5} & \underline{268.4} & \underline{91.9} & \underline{141.4} & 18.6 & 29.9 & 105.5 & 382.5 & 123.1 & 492.0  \\
\textbf{\textsc{Count2Density} (\textsc{man})}  & & \cmark  & &   \textbf{149.2} & \textbf{258.2} & \textbf{91.5} & \textbf{135.2} & 15.5 & 24.6 & \textbf{78.3} & \textbf{220.8}  & \textbf{122.2} & \textbf{485.3} \\

\bottomrule
\end{tabular}
\end{adjustbox}
\caption{Comparison of \textsc{Count2Density} with several related methodologies. For each approach, we report the type of annotation used: locations, counts, or self-labels. The \cmark~symbol indicates that the entire dataset is annotated (fully supervised); the $\blacktriangle$ symbol indicates that only 5-10\% of the dataset is annotated (semi-supervised). The * symbol indicates that the entire source domain is annotated (cross-domain). Best results are marked in \textbf{bold}, second best results are \underline{underlined}. Results are compared within each category (fully supervised, semi-supervised, cross-domain). Unless specified with a \textdagger~symbol, results are taken from the related publications.}
\label{tab:results}
\end{table*}

\subsection{Results}
\label{sec:results}

\begin{figure*}[t!]
\centering
\includegraphics[width=1\linewidth]{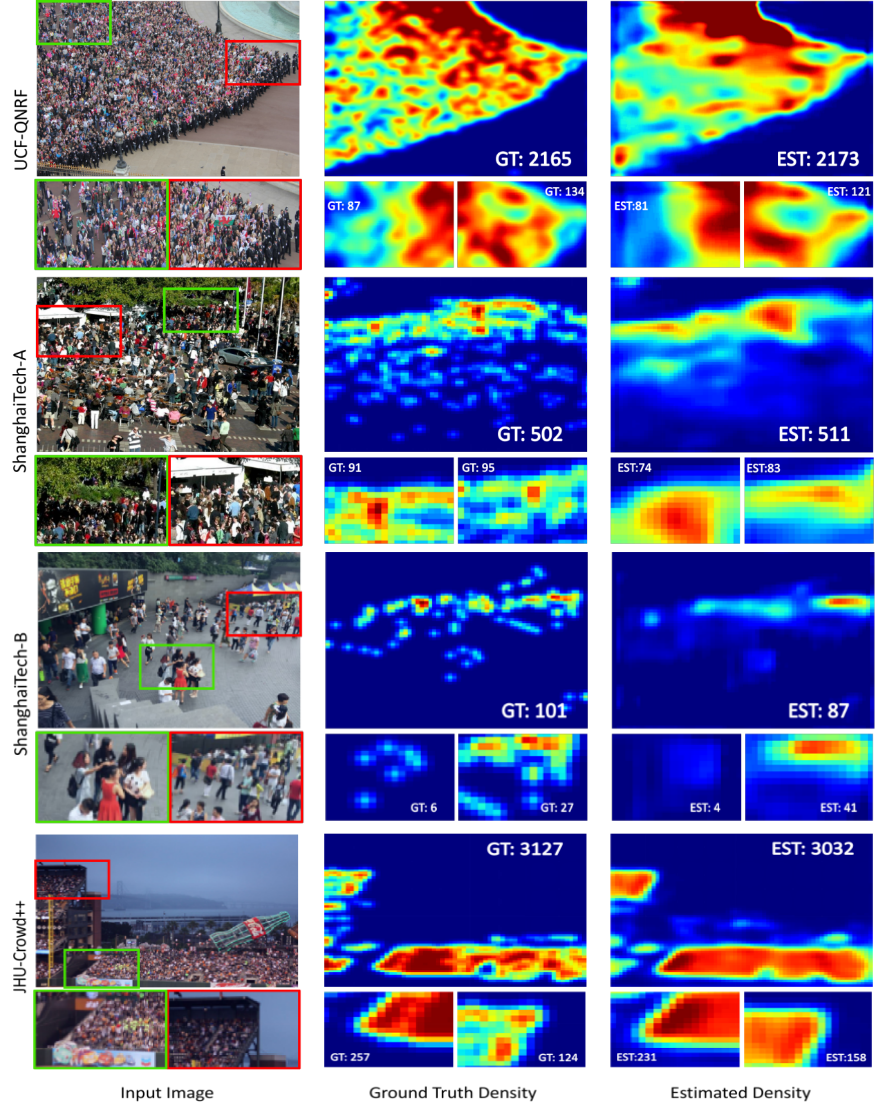}
\caption{Density maps predicted by \textsc{Count2Density} compared to ground-truth density, showing both global and subregion counting.}
\label{fig:density_maps}
\end{figure*}

\noindent \Cref{tab:results} presents the results of \textsc{Count2Density} compared with semi-supervised density estimation and cross-domain approaches. For completeness, we also report results obtained by fully supervised methodologies. Finally, \Cref{fig:density_maps} shows examples of estimated density maps.

We first compare \textsc{Count2Density} against a modified version of \textsc{irast} \cite{semisupervised2}, where we trained it using only count-level annotations to remove any contribution from location-level ones. In this scenario, the results in \Cref{tab:results} show that \textsc{irast} degrades in performance, yielding higher counting errors compared to our proposed method.

\mypar{Comparisons with Cross-Domain Approaches}
\Cref{tab:results} also presents comparisons with cross-domain approaches performing synthetic-to-real adaptation. Although these approaches use location-level annotations in the source domain, they are unsupervised when transferring to the target domain. Nevertheless, this still constitutes an advantage over \textsc{Count2Density}, which never sees location-level ground-truth annotations during training. Despite this, \textsc{Count2Density} drastically outperforms cross-domain approaches, especially on a large-scale dataset such as \textsc{ucf-qnrf}. Cross-domain approaches, however, yield better performance than ours on \textsc{ShanghaiTech-B}, as this dataset is characterised by less dense scenes. 
This occurs because the sampling strategy in \Cref{sec:unsupervised_noisy_spatial_prior} is better suited for highly dense scenes, as the sampled points are more likely to be situated within crowd regions.

\mypar{Comparisons with Semi-Supervised Approaches} \Cref{tab:results} shows comparisons with semi-supervised methods. Unlike \textsc{Count2Density}, these methods require a subset of the training set labelled with location-level annotations. Additionally, some semi-supervised approaches \cite{active_learning} augment the dataset by cropping images into patches, which have been shown to improve counting performance \cite{MAN,NCC,GL}. However, this training strategy cannot be applied in our method because we solely rely on count-level annotations, which are calculated over the entire image and thus lack count information on local patches.
Despite these major differences, \textsc{Count2Density} achieves comparable performance across all the datasets. Specifically, our method outperforms \textsc{irast} and \textsc{pal} on \textsc{jhu-crowd++}, reducing the \textsc{mae} by $\sim40\%$. On the \textsc{ShanghaiTech} datasets, \textsc{Count2Density} outperforms most of the semi-supervised approaches.

We also assess the performance of \textsc{Count2Density} when equipped with a few ground-truth location-level annotations and trained in a semi-supervised fashion. 
This is achieved by slightly modifying our pipeline as follows: for those images in the training set with location-level annotations, we provide the ground-truth $M_i$ instead of the pseudo-density map $\tilde{M}_i$ when evaluating $\mathcal{L}^{map}$ (\textit{c.f.} \Cref{eq:loss}). 
With this setup, \textsc{Count2Density} outperforms the recent \textsc{cu} on \textsc{ShanghaiTech-A} and \textsc{jhu-crowd++}, achieving the best \textsc{mae} of $70.5$ and $65.8$, respectively.

\begin{table}
    \centering
    \begin{adjustbox}{width=0.7\linewidth}
    \begin{tabular}{lcc|cc}
    \toprule
        Method & \textsc{mae} $\downarrow$ & \textsc{mse} $\downarrow$ & \textsc{ssim} $\uparrow$ & \textsc{psnr} $\uparrow$\\ 
    \midrule
         \textsc{cd-cc} \cite{Fua} & \underline{11.9} & \underline{18.9} & \underline{0.77} & \underline{21.6}\\
         \textsc{bla} \cite{BLA} & 12.8 & 20.6 & 0.72& 21.1 \\
         \textsc{irast} \cite{semisupervised2} & 27.5 & 31.8 & 0.58 & 18.75 \\
        \textbf{\textsc{count2density}} & \textbf{9.3} & \textbf{16.1} & \textbf{0.85} & \textbf{22.5} \\
    \bottomrule
    \end{tabular}
    \end{adjustbox}
\caption{Subregion counting performance and quality of predicted density maps on the \textsc{ucf-qnrf} dataset.}
    \label{tab:subregion_and_qualitative_results}

\end{table}

\mypar{Quality of Density Maps}
\Cref{fig:density_maps} shows qualitative examples of predictions made by \textsc{Count2Density} compared to the ground-truth. In \Cref{tab:subregion_and_qualitative_results}, we also quantitatively evaluate our predictions, assessing the ability to perform subregion counting. Specifically, for subregion counting, we divide the density maps into small tiles (subregions) and compare the counts between the predicted and ground-truth tiles. For quantitatively assess the quality of the density maps, we calculate the \textsc{ssim} and \textsc{psnr} between the predictions and ground truth. The results in \Cref{tab:subregion_and_qualitative_results} demonstrate that \textsc{Count2Density} outperforms competing methods in subregion counting. Additionally, our method achieves higher \textsc{ssim} and \textsc{psnr} values, indicating better quality density maps compared to other methods.

To further evaluate the quality of predicted density maps, we follow the methodology in \cite{steerer} to assess the localisation performance of \textsc{Count2Density}. Briefly, we determine location-level points from the predicted density maps by identifying local minima. \Cref{tab:localization_performance} shows that \textsc{Count2Density} outperforms \textsc{irast} and \textsc{adaseem} with a higher F1-score, and yields comparable performance with respect to a fully-supervised baseline, such as \textsc{steerer} \cite{steerer}.

\begin{table}
\centering
\begin{adjustbox}{width=0.7\linewidth}
        \begin{tabular}{lccc}
        \toprule
            Method & Precision $\uparrow$ & Recall $\uparrow$ & F1 $\uparrow$\\ 
        \midrule
            \textsc{steerer} (Full-sup.) \cite{steerer} & 78.6 & 72.7 & 75.5 \\
        \midrule
            \textsc{adaseem} \cite{adaseem} & \textbf{77.7} & 10.1 & 17.8\\
            \textsc{irast} \cite{semisupervised2} & 55.4 & 42.8 & 48.3\\
            \textbf{\textsc{count2density}} & 75.2 & \textbf{67.9} & \textbf{71.3} \\
        \bottomrule
        \end{tabular}
        \end{adjustbox}
\caption{Localisation performance on the UCF-QNRF dataset.}
\label{tab:localization_performance}
\end{table}

\subsection{Analysis}
\label{sec:analysis}

\mypar{Ablation Study} In \Cref{tab:ablation}, we report the results of the ablation study, by removing each individual component from our pipeline (\textit{c.f.} \Cref{fig:our_method}). Removing all components leads to the worst performance, with an \textsc{mae} of 218.2. By adding the Historical Map Bank, we boost performance reducing the \textsc{mae} by $\sim 30\%$. This result demonstrates that the Historical Map Bank can mitigate the effect of  confirmation bias, as discussed in \Cref{sec:unsupervised_noisy_spatial_prior}. Furthermore, the addition of either the saliency initialisation or the contrastive regularisation improves performance. Finally, it is clear that \textsc{Count2Density} achieves the best performance when all of three components work simultaneously. 

\begin{table}[t]

\centering
\begin{adjustbox}{width=0.7\linewidth}
\begin{tabular}{ccccc}
\toprule
    
\shortstack{Historical \\ Map Bank} & \shortstack{Saliency \\ Init.} & \shortstack{Contrastive \\ Reg.} & \textsc{mae} & \textsc{mse}
\\ \midrule
\textcolor{red}{\xmark} & \textcolor{red}{\xmark} & \textcolor{red}{\xmark} & 218.2 & 385.4
\\
\textcolor{mygreen}{\cmark} & \textcolor{red}{\xmark} & \textcolor{red}{\xmark} & 158.1 & 271.1
\\
\textcolor{red}{\xmark} & \textcolor{mygreen}{\cmark} & \textcolor{mygreen}{\cmark} & 206.4 & 354.2
\\
\textcolor{mygreen}{\cmark} & \textcolor{mygreen}{\cmark} & \textcolor{red}{\xmark} & 154.7 & 265.4
\\
\textcolor{mygreen}{\cmark} & \textcolor{red}{\xmark} & \textcolor{mygreen}{\cmark} & 152.5 & 259.0
\\
\textcolor{mygreen}{\cmark} & \textcolor{mygreen}{\cmark} & \textcolor{mygreen}{\cmark} & \textbf{149.2} &  \textbf{258.2}

\\ \bottomrule
\end{tabular}
\end{adjustbox}
\caption{\textsc{Count2Density} ablation analysis on \textsc{ucf-qnrf}.}
\label{tab:ablation}
\end{table}

\mypar{Hypergeometric \textit{vs.} Naive Sampling} To demonstrate the effectiveness of sampling from a hypergeometric distribution, as described in \Cref{sec:unsupervised_noisy_spatial_prior}, we assessed a naive sampling strategy by selecting the top-$y_i$ points from $\mathcal{H}_i^{(t)}$. On \textsc{ucf-qnrf}, the naive sampling strategy yields an \textsc{mae} of $200.1$, while our approach reduces the error by $\sim 25\%$ (our best \textsc{mae} on \textsc{ucf-qnrf} is 149.2, as reported in \Cref{tab:results}).

\begin{figure*}[t!]
    \centering
    \includegraphics[width=0.9\linewidth]{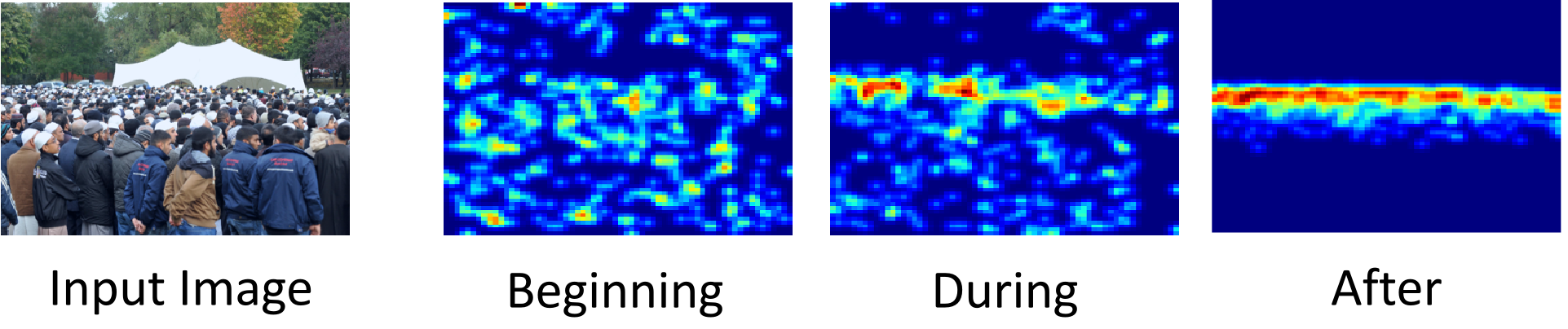}
    \caption{Pseudo-density maps generated during training. At the beginning, the pseudo-map is generated using the saliency estimator. During training, we apply \cref{eq:moving_avg} to refine the pseudo-map, refining the quantitative spatial information over time.}
    \label{fig:maps_refinement}
\end{figure*}

\mypar{Evolution of the Pseudo-Density Maps During Training} \Cref{fig:maps_refinement} shows the generated pseudo-density maps at different stages of training. At the first epoch, the pseudo-density maps are generated from the output provided by the unsupervised saliency estimator. Although this provides a spatial prior, the generated maps do not contain much spatial information at the beginning. During the training, our method progressively refines the pseudo-density maps, leading to meaningful estimations toward the end of the training. 

\mypar{Historical Map Bank Initialisation} We assess how much \textsc{Count2Density} depends on a good initialisation of the historical map bank to extract spatial information from count-level annotations. The ablation study in  \Cref{tab:ablation} shows that \textsc{Count2Density} performs well without initialising the historical map bank (\textsc{mae} is $\sim 2\%$ higher than when \textsc{bas-net} is used). We took this analysis a step further, and the results are in \Cref{tab:different_usod}. We initialised the historical map bank with a naive approach, by placing a blob with a fixed radius at the centre of each pseudo-density map. Although this strategy does not provide accurate spatial information, we observe a reduction in the error by $<1\%$ when compared with no initialisation. We also compared unsupervised \textsc{bas-net} against another unsupervised saliency estimator, \textsc{tgsd} \cite{newSOD}, showing that the unsupervised \textsc{bas-net} achieves the best performance. For completeness, we trained  \textsc{bas-net} in a supervised fashion to establish a lower bound  performance. Although it is expected that a supervised saliency estimator would improve performance,the such improvement is marginal, reducing the \textsc{mae} by $<0.5\%$, showing that \textsc{Count2Density} performs well without relying on location-level annotations to initialise $\mathcal{H}$. Overall, it can be observed in \Cref{tab:different_usod} that \textsc{Count2Density} is not overly reliant on the initialisation of the historical map bank and the ablation in \Cref{tab:ablation}. Despite  the performance reduction observed with the historical map bank, best results are achieved when the moving average in \cref{eq:moving_avg} is used in tandem with the contrastive regulariser and saliency initialisation with the unsupervised \textsc{bas-net} \cite{basnet}.

\mypar{Computational efficiency} We train \textsc{Count2Density} for 18-24 hours on a single NVIDIA RTX A6000 depending on the backbone used. We observed a peak of about 10GB of GPU ram usage during training and 4GB during inference. For inference, \textsc{Count2Density} takes about $0.3s$ to process a single image and provide the corresponding density map.

\section{Conclusions and Limitations} \noindent 

\begin{table}[t]

\centering
\begin{tabular}{lcc}
\toprule
Method & \textsc{mae} & \textsc{mse}
\\ \midrule
\textit{None} & 152.5 & 259.0 \\
Centred blob & 151.4 & 260.4 \\
\textsc{tgsd} \cite{newSOD} & 150.1 & \textbf{257.9} \\
Unsup. \textsc{bas-net} \cite{basnet}~ & \textbf{149.2} & 258.2 \\
\midrule
Sup. \textsc{bas-net} \cite{basnet} & 148.6 & 257.4  
\\ \bottomrule
\end{tabular}
\caption{Comparison with different saliency estimators used to initialise the historical map bank.}
\label{tab:different_usod}
\end{table}

To reduce the cost of collecting fine-grained annotation for crowd density estimation, we introduced \textsc{Count2Density}, a method that leverages only count-level annotations to predict meaningful density maps, also enabling subregion counting. Our approach is capable of extracting spatial information from count values to generate pseudo-density maps, which are then used to train the model in a self-supervised fashion. We show that \textsc{Count2Density} outperforms cross-domain approaches by a large margin across several datasets and shows significant improvements in performance when trained in a semi-supervised setting, surpassing the state-of-the-art.

While our method demonstrates strong performance, it is not without limitations. One notable challenge is the memory demands required to store the historical map bank, which may limit the scalability of our approach for larger datasets or real-time applications. Additionally, the training time and computational overhead associated with maintaining the map bank could be prohibitive in certain settings. Future work should focus on more efficient memory management strategies or alternative architectural designs to address these concerns.

Despite these limitations, our method presents a promising direction for crowd density estimation and other domains where spatial information is unavailable during training. \textsc{Count2Density} opens up opportunities in challenging scenarios where location-level annotations are scarce or impractical to collect. Overall, our approach offers an effective solution in density estimation, with potential applications far beyond crowd counting.



\bibliographystyle{elsarticle-num-names} 
\bibliography{main.bib}
\end{document}